\documentclass[letterpaper]{article} % DO NOT CHANGE THIS
\usepackage{aaai25}  % DO NOT CHANGE THIS
\usepackage{times}  % DO NOT CHANGE THIS
\usepackage{helvet}  % DO NOT CHANGE THIS
\usepackage{courier}  % DO NOT CHANGE THIS
\usepackage[hyphens]{url}  % DO NOT CHANGE THIS
\usepackage{graphicx} % DO NOT CHANGE THIS
\usepackage{natbib}  % DO NOT CHANGE THIS AND DO NOT ADD ANY OPTIONS TO IT
\usepackage{caption} % DO NOT CHANGE THIS AND DO NOT ADD ANY OPTIONS TO IT
\usepackage{algorithm}
\usepackage{algorithmic}
\usepackage{amsmath}

\usepackage{newfloat}
\usepackage{listings}
\DeclareCaptionStyle{ruled}{labelfont=normalfont,labelsep=colon,strut=off} % DO NOT CHANGE THIS
\floatstyle{ruled}
\newfloat{listing}{tb}{lst}{}
\floatname{listing}{Listing}
\usepackage{colortbl}
\usepackage{float}
\usepackage{placeins}
\title{Object Gaussian for Monocular 6D Pose Estimation from Sparse Views}
\author{
    Luqing Luo\equalcontrib\textsuperscript{\rm 1},
    Shichu Sun\equalcontrib\textsuperscript{\rm 1},
    Jiangang Yang\textsuperscript{\rm 1},
    Linfang Zheng\textsuperscript{\rm 2},
    Jinwei Du\textsuperscript{\rm 3},
    Jian Liu$^{\dagger}$\textsuperscript{\rm 1}
}
\affiliations{
    \textsuperscript{\rm 1}Institute of Microelectronics Chinese Academy of Sciences, Beijing, China\\
    \textsuperscript{\rm 2}University of Birmingham, Birmingham, UK\\
    \textsuperscript{\rm 3}NVIDIA, Shanghai, China
}
\usepackage{bibentry}
\begin{document}

\maketitle

\begin{abstract}
Monocular object pose estimation, as a pivotal task  in computer vision and robotics, heavily depends on accurate 2D-3D correspondences, which often demand costly CAD models that may not be readily available. Object 3D reconstruction methods offer an alternative, among which recent advancements in 3D Gaussian Splatting (3DGS) afford a compelling potential. Yet its performance still suffers and tends to overfit with fewer input views. Embracing this challenge, we introduce SGPose, a novel framework for sparse view object pose estimation using Gaussian-based methods. Given as few as ten views, SGPose generates a geometric-aware representation by starting with a random cuboid initialization, eschewing reliance on Structure-from-Motion (SfM) pipeline-derived geometry as required by traditional 3DGS methods. SGPose removes the dependence on CAD models by regressing dense 2D-3D correspondences between images and the reconstructed model from sparse input and random initialization, while the geometric-consistent depth supervision and online synthetic view warping are key to the success. Experiments on typical benchmarks, especially on the Occlusion LM-O dataset, demonstrate that SGPose outperforms existing methods even under sparse view constraints, under-scoring its potential in real-world applications. 

\end{abstract}


\section{Introduction}

Monocular pose estimation in 3D space, while inherently ill-posed, is a necessary step for many tasks involving human-object interactions, such as robotic grasping and planning~\cite{azad2007stereo}, augmented reality~\cite{tan2018real}, and autonomous driving~\cite{manhardt2019roi, qi2018frustum}. 
Influenced by deep learning approaches, its evolution has enabled impressive performance even in cluttered environments.  
The most studied task in this field assumes that the CAD model of the object is known a priori~\cite{peng2019pvnet, li2019cdpn, park2019pix2pose, cai2020reconstruct, chen2020category, park2020latentfusion}, but the accessibility of such a predefined geometry information prevents its applicability in real-world settings. 
To reduce reliance on specific object CAD models, recent research has shifted toward category-level pose estimation~\cite{wang2019normalized, ahmadyan2021objectron}, aiming to generalize across objects within the same category. However, these methods typically ask for extra depth information, and can falter with instances of varying appearances.

\begin{figure}[t]
\centering
\includegraphics[width=0.47\textwidth]{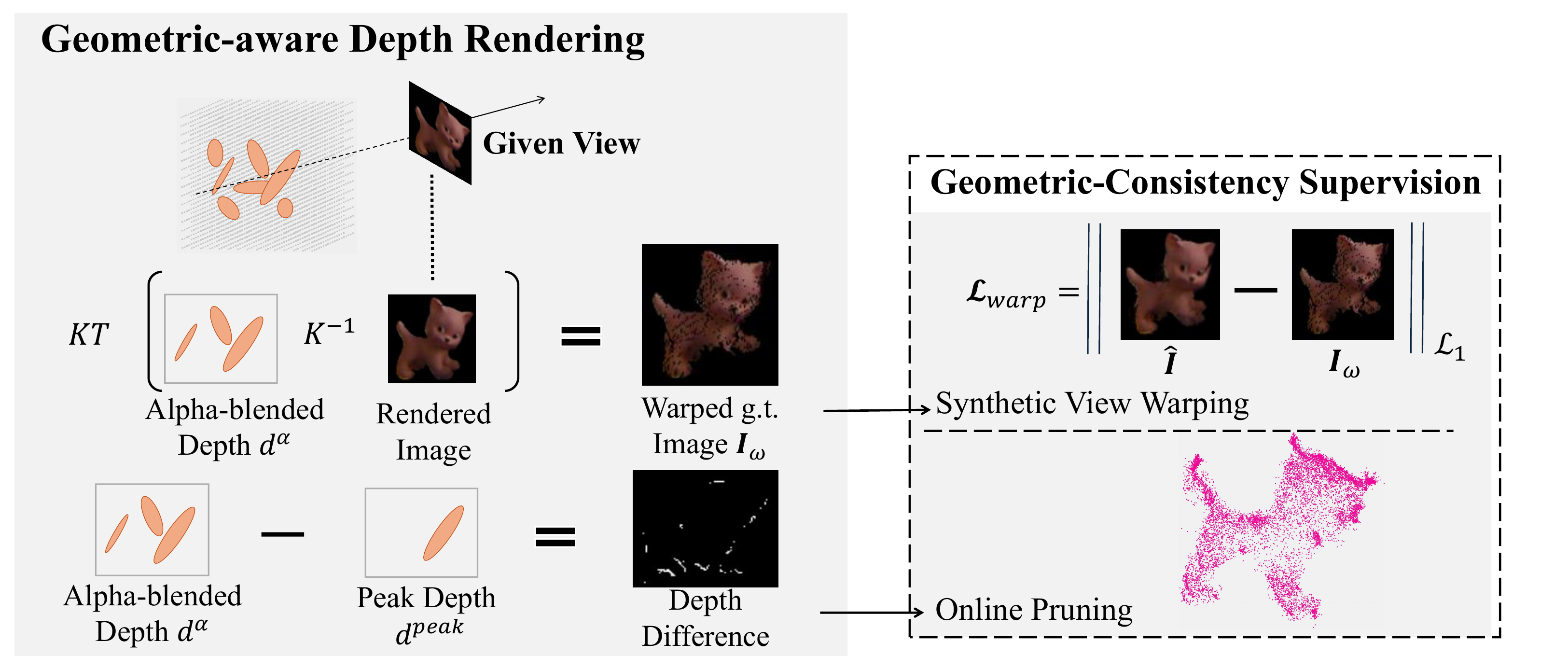} % Reduce the figure size so that it is slightly narrower than the column.
\caption{The alpha-blended depth $d^{\alpha}$ integrates depth across Gaussian primitives along the ray, the peak depth $d^{peak}$ selects the one of highest opacity.
$d^{\alpha}$ enables reliable online synthetic view warping without leveraging external depth information, in conjunction with $d^{peak}$ guides the online pruning, both of them contribute to object Gaussian reconstruction under sparse views.}
\label{fig0}
\end{figure}

Emerging real-world demands call for an object pose estimator generalizable, flexible, and computationally efficient. 
Ideally, a new object can be reconstructed from casually taken reference images, sans the need for fine-grained well-textured 3D structures.
Reconstruction-based methods have shown the feasibility of this proposal~\cite{cai2020reconstruct, liu2022gen6d, sun2022onepose, he2022onepose++, li2023nerf, cai2024gs}, which basically reconstruct the 3D object from the multi-view RGB images to substitute missing CAD model.
however, reconstruction-based methods have long relied on a fixed budget of high-quality given images and the prerequisite use of Structure-from-Motion (SfM) techniques, resulting in a notably tedious and costly training process.
Our method deviates from these requirements by pioneering an efficient object reconstruction method that thrives on limited reference images and the convenience of random initialization.
Capitalizing on the high-quality scene representation and real-time rendering of 3DGS~\cite{kerbl20233d}, we unveil SGPose, an novel framework of Sparse View Object Gaussian for monocular 6D Pose estimation. 
The proposed SGPose develops geometric-aware depth to guide object-centric 3D representations from RGB only input. Requiring a mere ten views, SGPose achieves competitive performance for object pose estimation, heralding its readiness for real-world deployment.

In our work, we extend a variant of 3DGS~\cite{huang20242d} by formulating Gaussian primitives as elliptic disks instead of ellipsoids to derive depth rendering.
As illustrated in Fig.~\ref{fig0}, the conceived geometric-consistent constraints guide depth acquisition.
The resulting depth enables reliable online synthetic view warping and eases the challenges of sparse views for traditional 3DGS methods.
Additionally, an online pruning is incorporated in terms of the geometric-consistent depth supervision, toning down common sparse view reconstruction issues like floaters and background collapses. 
In contrast to the conventional reliance on SfM pipelines for point cloud initialization in 3DGS, the proposed SGPose opts for a random initialization from a cuboid of 4,096 points. 
The proposed object Gaussian generates dense image pixels and object coordinates correspondence (2D-3D correspondence) maps efficiently using geometric-aware depth rendering, serving as a keystone advantage for monocular 6D pose estimation. An adapted GDRNet++ framework~\cite{liu2022gdrnpp_bop} is utilized to assess 6D pose estimation on the LM~\cite{hinterstoisser2012model} and Occlusion LM-O~\cite{brachmann2014learning} datasets. 
Our SGPose takes sparse view images and pose annotations to create synthetic views, object masks, and dense correspondence maps.
Noteworthy, for the Occlusion LM-O dataset, we render data similar to PBR (Physically Based Rendering) data~\cite{Denninger2023}, which further enhance the performance of the proposed method.
By matching state-of-the-art performance across CAD-based and CAD-free approaches, we highlight the efficiency and flexibility of our method. 
To sum up, Our main contributions are:
\begin{itemize}
\item By intaking only RGB images, the proposed geometric-aware object Gaussian derives accurate depth rendering from random point initialization;
\item The rendered depth ensures a reliable synthetic view warping and an effective online pruning, addressing the issue of overfitting under sparse views at an impressively low time cost;
\item By generating dense 2D-3D correspondences and images that simulate real occlusions using the proposed object Gaussian, our SGPose framework achieves CAD-free monocular pose estimation that is both efficient and robust. 
\end{itemize}

\section{Related Work}

\subsubsection{CAD-Based Object Pose Estimation}

Many previous works on pose estimation rely on known CAD models. Regression-based methods~\cite{kehl2017ssd, labbe2020cosypose, li2018deepim, xiang2017posecnn} estimate pose parameters directly from features in regions of interest (RoIs), while keypoint-based methods establish correspondences between 2D image pixels and 3D object coordinates either by regression~\cite{oberweger2018making, park2019pix2pose, pavlakos20176} or by voting~\cite{peng2019pvnet}, often solve poses by using a variant of Perspective-n-Points (PnP) algorithms~\cite{lepetit2009epnp}.NOCS~\cite{wang2019normalized} establishes correspondences between image pixels and Normalized Object Coordinates (NOCS) shared across a category, reducing dependency on CAD models at test time. Later works~\cite{lee2021category,tian2020shape, wang2021gdr, wang2021category} build upon this idea by leveraging category-level priors to recover more accurate shapes.
A limitation of these methods is that objects within the same category can have significant variations in shape and appearance, which challenges the generalization of trained networks. Additionally, accurate CAD models are required for generating ground-truth NOCS maps during training. In contrast, our framework reconstructs 3D object models from pose-annotated images, enabling CAD-free object pose estimation during both training and testing phases.

\subsubsection{CAD-Free Object Pose Estimation}

Some endeavors have been made to relax the constraints of CAD models of the objects. 
RLLG~\cite{cai2020reconstruct} uses multi-view consistency to supervise coordinate prediction by minimizing reprojection error.  
NeRF-Pose~\cite{li2023nerf} trains a NeRF-based~\cite{mildenhall2021nerf} implicit neural representation of object and regresses object coordinate for pose estimation.
Gen6D~\cite{liu2022gen6d} initializes poses using detection and retrieval but requires accurate 2D bounding boxes and struggles with occlusions. 
GS-Pose~\cite{cai2024gs} improves on Gen6D~\cite{liu2022gen6d} by employing a joint segmentation method and 3DGS-based refinement. 
OnePose~\cite{sun2022onepose} reconstructs sparse point clouds of objects and extracts 2D-3D correspondences, though its performance is limited on symmetric or textureless objects due to its reliance on repeatable keypoint detection.
While OnePose++~\cite{he2022onepose++} removes the dependency on keypoints resulting in a performance enhancement.
Unlike these methods, which require numerous input images for training, we directly leverage the power of 3DGS~\cite{kerbl20233d} for geometric-aware object reconstruction from sparse, pose-annotated images to achieve pose estimation.

\section{Methods}

\begin{figure*}[t]
\centering
\includegraphics[width=0.9\textwidth]{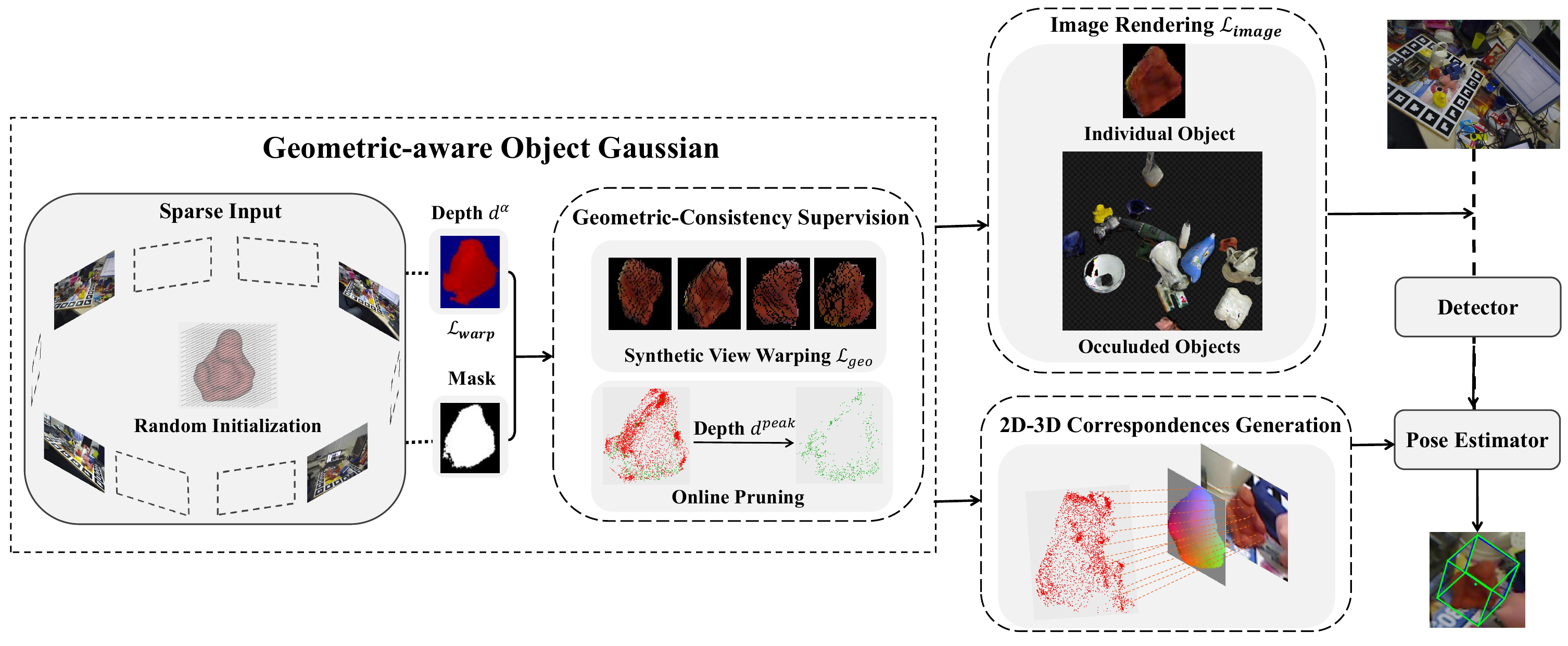} % Reduce the figure size so that it is slightly narrower than the column.
\caption{SGPose Pipeline. Given sparse RGB images and a cuboid random initialization, 
the object Gaussian learns the geometry of target objects under the supervision of geometric-consistency, to render synthetic views, including both of individual object images and occluded objects images, masks and dense 2D-3D correspondences.  The image rendering loss $\mathcal{L}_{image}$, image warping loss $\mathcal{L}_{warp}$ and geometric-consistent loss $\mathcal{L}_{geo}$ are used to guide the learning process.
For pose estimation, the objects are detected and cropped from test images by detector~\cite{redmon2018yolov3}, the above rendering results, as the replacement of CAD models, are feed to pose estimator~\cite{liu2022gdrnpp_bop} for regression.}
\label{fig1}
\end{figure*}

An overview of the proposed method is presented in Fig.~\ref{fig1}.
Given sparse views as input, the dense 2D-3D correspondence maps are encoded in the conceived object Gaussian naturally, by supervising geometric-aware depth. Consequently, the created synthetic views and correspondence maps are availed to a downstream pose estimator, to achieve CAD-free monocular Pose Estimation.

\subsection{Depth Rendering of Geometric-aware Object Gaussian}

The object geometry is described by Gaussian primitives of probability density function as~\cite{kerbl20233d}, 
\begin{equation}
\mathcal{G}(\mathbf{x})=e^{-\frac{1}{2}(\mathbf{x}-\mathbf{\mu})^{\top} \Sigma^{-1}(\mathbf{x}-\mathbf{\mu})},
\end{equation}
\noindent where $\mathbf{x}$ is a point in world space to describe the target object and $\mu$ is the mean of each Gaussian primitive (which also is the geometric center). Thus, the difference vector $\mathbf{x}-\mu$ indicts the probability density of $\mathbf{x}$, which peaks at the center $\mu$ and decreases as departing from it. 

By treating Gaussian primitives as the elliptical disks~\cite{huang20242d}, the covariance matrix $\Sigma$ is parameterized on a local tangent plane centered at $\mu$ with a rotation matrix and a scaling matrix. 
Concretely, the rotation matrix $R$ is comprised of three vectors $\mathbf{t}_u, \mathbf{t}_v$ and $\mathbf{t}_w$, where two orthogonal tangential vectors $\mathbf{t}_u$ and $\mathbf{t}_v$ indicate the orientations within the local tangent plane, and $\mathbf{t}_w=\mathbf{t}_u \times \mathbf{t}_v$ represents the normal perpendicular to the plane. 
The scaling matrix $S$ depicts the variances of Gaussian primitives on corresponding directions, noted that there is no distribution in the direction of $\mathbf{t}_w$ since the Gaussian primitive is defined on a flat elliptical disk. 
Thereby, the $3 \times 3$ rotation matrix $R=\left[\mathbf{t}_u, \mathbf{t}_v, \mathbf{t}_w\right]$ and the scaling matrix $S=\left[s_u, s_v, 0\right]$ form up the covariance matrix as $\Sigma = RS{S}^{\top}{R}^{\top}$.

By leveraging the world-to-camera transformation matrix $W$ and the Jacobian of the affine approximation of the projective transformation matrix $J$, the projected 2D covariance matrix $\Sigma^{\prime}$ in camera coordinates is given as, $\Sigma^{\prime}=J W \Sigma W^{\top} J^{\top}$.
By virtue of the same structure and properties are maintained by skipping the third row and column of $\Sigma^{\prime}$ ~\cite{kopanas2021point, zwicker2001ewa}, a $2 \times 2$ variance matrix $\Sigma^{2 D}$ (corresponding to $\mathcal{G}^{2 D}$) is obtained,
\begin{equation}
\mathcal{G}^{2 D}(\mathbf{x}^{\prime}) =  e^{-\frac{1}{2}\left(\mathbf{x}^{\prime}-\mathbf{\mu}^{\prime}\right)^{\top} \left({\Sigma}^{2 D}\right)^{-1} \left(\mathbf{x}^{\prime}-\mathbf{\mu}^{\prime}\right)},
\end{equation}
\noindent where $\mathbf{x}^{\prime}$ and $\mu^{\prime}$ stands for the projected points of  $\mathbf{x}$ and $\mu$ in the screen space, respectively.

Furthermore, the local tangent plane is defined as,
\begin{equation}
X(u, v)=\mu+s_u \mathbf{t}_u u+s_v \mathbf{t}_v v=\mathbf{H}(u, v, 1,1)^{\top},
\end{equation}
\noindent where $\mathbf{H}=\left[\begin{array}{cccc}s_u \mathbf{t}_u & s_v \mathbf{t}_v & \mathbf{0} & \mu \\ 0 & 0 & 0 & 1\end{array}\right]=\left[\begin{array}{cc}RS & \mu \\ \mathbf{0} & 1\end{array}\right]$ is a homogeneous transformation matrix mapping point $(u,v)$ on local tangent plane into the world space. 

Suppose there is a ray emitting from the camera optical center onto the screen space. 
The geometric-aware depth $d^\text{geo}$ is hence defined as the distance between the camera and the Gaussian primitive along the ray. Accordingly, the homogeneous coordinate of the point $(u,v)$ projected onto the screen is~\cite{zwicker2004perspective}, 
\begin{equation}\label{eq3}
({u^{\prime}}, {v^{\prime}}, d, w)^{\top}=W X(u, v)=W \mathbf{H}(u, v, 1,1)^{\top}.
\end{equation}
\noindent where $w$ is usually set to 1 (the homogeneous representation describes a point when $w \neq 0$, while it depicts a ray when $w = 0$). 
This point can be further represented as the intersection of two orthogonal planes corresponding to $u^{\prime}$ and $v^{\prime}$~\cite{weyrich2007hardware, sigg2006gpu}.
Specifically , $u^{\prime}$-plane is defined by a normal vector $(-1, 0, 0)$ and an offset $u^{\prime}$, the 4D homogeneous plane thus is $\mathbf{h}_{u^{\prime}}=(-1, 0, 0, u^{\prime})$. Similarly, $v^{\prime}$-plane is $\mathbf{h}_{v^{\prime}}=(0, -1, 0, v^{\prime})$. Conversely, both planes can be transformed back to the local tangent plane coordinates as,
\begin{equation}
\mathbf{h}_u=(W\mathbf{H})^{\top} \mathbf{h}_{u^{\prime}}, \quad \mathbf{h}_v=(W\mathbf{H})^{\top} \mathbf{h}_{v^{\prime}},
\end{equation}
\noindent in which $(W\mathbf{H})^{\top}$ is equivalent to $(W\mathbf{H})^{-1}$ as show in~\cite{vince2008geometric}.
According to~\cite{huang20242d}, since the screen point $(u^{\prime}, v^{\prime})$ must lie on both $u^{\prime}$-plane and $v^{\prime}$-plane, for any point $(u, v, 1, 1)$ on the elliptical disk, the dot product of the transformed plane $\mathbf{h}_u$ and $\mathbf{h}_v$ with the point \((u, v, 1, 1)\)  should be zero,
\begin{equation}
\mathbf{h}_u \cdot(u, v, 1,1)^{\top}=\mathbf{h}_v \cdot(u, v, 1,1)^{\top}=0,
\end{equation}
\noindent by solving the equation above, the coordinates of the screen point $(u^{\prime}, v^{\prime})$ on the local tangent plane are yielded,
\begin{equation}
u=\frac{\mathbf{h}_u^2 \mathbf{h}_v^4-\mathbf{h}_u^4 \mathbf{h}_v^2}{\mathbf{h}_u^1 \mathbf{h}_v^2-\mathbf{h}_u^2 \mathbf{h}_v^1}, \quad v=\frac{\mathbf{h}_u^4 \mathbf{h}_v^1-\mathbf{h}_u^1 \mathbf{h}_v^4}{\mathbf{h}_u^1 \mathbf{h}_v^2-\mathbf{h}_u^2 \mathbf{h}_v^1},
\end{equation}
\noindent where $\mathbf{h}_u^i, \mathbf{h}_v^i$ are the elements of the 4D homogeneous plane parameters.

Thus far, the Gaussian primitives can be expressed with respect to $(u,v)$. Suppose $\Sigma^{2 D} M=I$, by transforming the 2D covariance matrix $\Sigma^{2 D}$ into the identity matrix $I$, the probability density function can be rewritten as standardized Gaussian (with mean of zero and deviation of one),
\begin{equation}
\mathcal{G}(\mathbf{x}^{\prime})=e^{-\frac{1}{2}\left(M\left(\mathbf{x}^{\prime}-\mu^{\prime}\right)\right)^{\top}\left(M\left(\mathbf{x}^{\prime}-\mu^{\prime}\right)\right)},
\end{equation}
\noindent where $\mathbf{x}^{\prime}$ can be further replaced by $(u, v)$ via some linear transformations as,
\begin{equation}
\mathcal{G}(u, v)=e^ {-\frac{1}{2}(u^2+v^2)}.
\end{equation}
To further take account of numerical instability introduced by inverse homogeneous transformations of Eq.~\ref{eq3},  a lower bounded Gaussian is imposed~\cite{botsch2005high}, 
\begin{equation}
\hat{\mathcal{G}}(u,v)=\max \left\{\mathcal{G}(u,v), \mathcal{G}\left(\frac{(u^{\prime},v^{\prime})-\mu^{\prime}}{r}\right)\right\}.
\end{equation}
\noindent When the elliptic disk is projected as the segment line in some cases, a low-pass filter (centered at $\mu^{\prime}$ with radius $r$) is wielded to guarantee sufficient points passed toward the screen space (the radius is set as $\sqrt{2} / 2$ empirically by following~\cite{huang20242d}).

Suppose the opacity of $i$-th Gaussian primitive is $\alpha_i$, by considering the alpha-weighted contribution along the ray, the accumulated transmittance is,
\begin{equation}
T_i=\prod_{j=1}^{i-1}\left(1-\alpha_j {\hat{\mathcal{G}_j}(u, v)}\right).
\end{equation}
To this end, the proposed object Gaussian renders both image and depth map of the object. The final color is $c^{\alpha}=\sum_{i \in \mathcal{N}}T_i \alpha_i \hat{\mathcal{G}_i}(u, v) c_i $, with $c_i$ as the view-dependent appearance represented by spherical harmonics  (SH)~\cite{fridovich2022plenoxels, takikawa2022variable}.
The alpha-blended depth map is formulated via the summation of geometric-aware depth $d^\text{geo}$ of each Gaussian primitive as,
\begin{equation}
d^{\alpha}=\sum_{i \in \mathcal{N}}T_i \alpha_i \hat{\mathcal{G}_i}(u, v) \max \left\{d^\text{geo}_i \mid T_i>\sigma\right\},
\end{equation}
\noindent where $\sigma=0.5$ is a threshold deciding whether the rendered depth valid. Noted that the maximum depth along the ray is selected if $T_i$ does not reach the threshold as in~\cite{huang20242d}. 

\subsection{Geometric-Consistency under Sparse Views}

In the circumstance of extremely sparse view reconstruction, the object Gaussian struggles with over-fitting~\cite{jain2021putting,niemeyer2022regnerf}, where the background collapse and floaters are commonly witnessed even the rendered view deviates marginally from the given one~\cite{xiong2023sparsegs}. 
% In particular, background collapse stems from distant surface modeled incorrectly as close translucent dense clouds. 
% Floaters are discontinuous regions of high density that appear as fuzzy clouds in certain views.
In principle, the effective solutions involve online synthetic view augmentation and geometric-consistent depth supervision.
Notably, effective online synthetic view augmentation remarkably reduces the need of a high budget of real images, and the multi-view geometric consistency prevents significant fluctuations on rendered depth.

\paragraph{Synthetic View Warping}
Given what is at stake, it is intuitive to warp synthetic views online to enrich training samples, which encourages model to adapt from a diverse set of perspectives and brings better generalization capabilities on unseen views. 

Owing to lack of ground truth of synthetic views, the proposed alpha-blended depth map plays an essential role in warping process. 
Specifically, the rendered depth $d^{\alpha}$ is used to transform the given view into 3D points, which are re-projected as pixels of synthetic views. 
Formally, the pixel $(u^{\prime}_{g},v^{\prime}_{g})$ of a given view is warped as $(u^{\prime}_{w},v^{\prime}_{w})$ of an unseen view, which is
\begin{equation}
\left(u_w^{\prime}, v_w^{\prime}\right)=K T\left[d^\alpha K^{-1}\left(u_g^{\prime}, v_g^{\prime}, 1\right)^{\top}\right],
\end{equation}
\noindent where $K$ is the camera intrinsic, and the rendered depth $d^\alpha$ serves as the pixel-wise scaling factor to confine the re-projection within a meaningful range.

Moreover, the transformation $T$ from give view to warped view is obtained via perturbations (including rotations and translations) sampled from normal distribution randomly, by making use of tool provided in~\cite{li2018deepim}.
The warped pixels are assembled as the ground truth image $I_{w}$,
\begin{equation}
\mathcal{L}_{warp}=\mathcal{L}_1 (\hat{I}, \;  I_{w}).
\end{equation}
\noindent While the ground truth image $I_{w}$ and the rendered image $\hat{I}$ of a specific synthetic view establish supervision of the image warping loss via $\mathcal{L}_1$.

\paragraph{Depth Supervision under Geometric-Consistent}
Considering each Gaussian is in tandem with the depth distribution of a certain region in the scene, to concentrate the geometric-aware depth of Gaussian primitives along the ray is beneficial to refine each Gaussian's contribution to overall distribution. 
Accordingly, the geometric-consistent loss is employed as
\begin{equation}
\mathcal{L}_{geo}=\sum_{i, j} \omega_i \omega_j\left|d^\text{geo}_i-d^\text{geo}_j\right|,
\end{equation}
\noindent where $\omega_i= T_i \alpha_i \hat{\mathcal{G}_i}(u, v)$ is the blending weight of $i$-th Gaussian primitive~\cite{huang20242d}.

\paragraph{Geometric-consistency Guided Online Pruning}
Lastly, inspired by~\cite{xiong2023sparsegs}, an online floaters pruning strategy is implemented by introducing a peak depth,
\begin{equation}
d^{peak}= d^\text{geo}_{\arg \max _i\left(\omega_i\right)}.
\end{equation}
\noindent The peak depth~\cite{xiong2023sparsegs} is acquired by selecting the Gaussian of highest blending weight, which also is the Gaussian of the highest opacity.

To implement the multi-view geometric-aware depth comparison, the alpha-blending depth $d^{\alpha}$ and peak depth $d^{peak}$ are compared under each given view.
Generally, the alpha-blending depth locates slightly behind the peak depth, the differences result in a candidate region for pruning. 
While the corresponding opacity $\alpha_i$ of peak depth within the region guides the online floater pruning. 

\subsection{2D-3D Correspondence Generation}

The proposed SGPose starts from RGB data alone, without taking advantage of external depth information, yet it effectively renders reliable geometric-aware depth maps.
Unlike traditional CAD-free pipelines that heavily rely on geometric initialization from SfM methods such as COLMAP~\cite{schoenberger2016mvs, schoenberger2016sfm}, our method handles the random initialization of a cuboid that approximates the bounding box of object. 
The differentiable optimization of the proposed object Gaussian is expressed as
\begin{equation}
\mathcal{L}=\mathcal{L}_{image}+\mathcal{L}_{warp}+\lambda_1 \mathcal{L}_{geo}+\lambda_2 \mathcal{L}_{normal}.
\end{equation}
\noindent Concretely, $\mathcal{L}_{image}$ is the image rendering loss combining $\mathcal{L}_1$ with the D-SSIM term from~\cite{kerbl20233d}, which is implemented in the given views only.
$\mathcal{L}_{image}$ for given views and $\mathcal{L}_{warp}$ for synthetic views follow the identical optimization pipeline, which update the object Gaussian model alternatively. 
The reasons of implementing such a training strategy are two-folded. 
Firstly, data from respective views exhibit disparate geometric details, tackling with them independently accommodates the model to the diversified data distributions; Secondly, the stand-alone Gaussian densification and pruning mitigate fluctuations brought by view alternating. 
$\lambda_1$ is set as $10^4$ to align up the scale of depth term with the other ones, and $\lambda_2=0.005$ for normal loss $\mathcal{L}_{normal}=\sum_i \omega_i\left(1-\mathrm{n}_i^{\top} \phi(u,v)\right)$ to facilitate the gradients of depth maps $\phi(u,v)$ in line with normal maps $\mathrm{n}_i$~\cite{huang20242d}.

Overall, the geometric-consistent supervision under sparse views reconstructs the desirable object Gaussian.
The dense 2D-3D correspondences, generated to fully replace the CAD models, along with synthetic view color images and object masks, serve as the ground truth for a modified GDRNet++~\cite{liu2022gdrnpp_bop} to perform monocular pose regression. 
Among them, the generation of 2D-3D correspondences and the simulation of realistic occlusions in images are essential for the task.

For dense 2D-3D correspondences, object points are obtained by transforming the rendered depth map into 3D points of camera coordinates via the known camera's intrinsic, and in turn mapping the points to world space via the specific view parameters (rotations and translations). Pixel coordinates are calculated from the rendered object mask.
Thus, the 3D points in world space and corresponding pixel coordinates are stack orderly as dense 2D-3D correspondences of any specific view, which is
\begin{equation}
\mathbf{M}_{2D-3D}=
\left[\begin{array}{c}
\mathbf{R}_{obj}^{\top}\left(d^\alpha K^{-1}(u^{\prime}, v^{\prime}, 1)^{\top}-\mathbf{t}_{obj}\right) \\
(u^{\prime}, v^{\prime})_{mask}
\end{array}\right],
\end{equation}
\noindent where $\mathbf{R}_{obj}$ and $\mathbf{t}_{obj}$ are the specific view parameters. 

\begin{table*}[t]
\centering
\resizebox{0.92\textwidth}{!}{
\begin{tabular}{c|ccccc|ccccccc|c}
\hline
Object & DPOD 
& PVNet& CDPN & GDR-Net & SO-Pose & RLLG &Gen6D$^{\dagger}$&One& One&  GS-Pose & NeRF-&NeRF-& \textbf{Ours} \\ 
& & &  & & & & & Pose & Pose++& & Pose& Pose$^{\dagger}$ \\
\hline
Views & &&&&& $\sim$200&$\sim$200&& $\sim$180& $\sim$180  & 156 & 156 & \bf10 \\\hline
CAD & \multicolumn{5}{c|}{\textit{w/} CAD}& \multicolumn{8}{c}{\textit{w/o} CAD} \\ \hline
Ape & 87.73  & 43.6& 67.33 & 76.29 & 85.43 & 52.9&-&11.8& 31.2&  65.1& 89.1&93.1 & 82.57  \\
Bvise& 98.45 & 99.9& 98.74 & 97.96 & 99.42 & 96.5&77.03&92.6& 97.3&  95.7&  99.3&99.6 & 99.32 \\
Camera & 96.07 & 86.9& 92.84 & 95.29 & 96.67 & 87.8&66.67&88.1& 88.0&  89.4& 98.7&98.9 & 96.18 \\
Can & 99.71 &  95.5& 96.56 & 98.03 & 98.62 & 86.8&-&77.2& 89.8&  97.2& 99.1&99.7 & \bf\textit{99.11} \\
Cat & 94.71 & 79.3& 86.63 & 93.21 & 95.01 & 67.3&60.68&47.9& 70.4&  84.6& 97.1&98.1 & \bf\textit{95.71} \\
Driller & 98.80 & 96.4& 95.14 & 97.72 & 98.41 &  88.7&67.39&74.5&  92.5&  90.7&97.4&98.7 & \bf 98.91 \\
Duck & 86.29 & 52.6 & 75.21 & 80.28 & 85.73 & 54.7&40.47&34.2& 42.3&  72.3&90.3&94.2 & 85.26 \\
Eggbox& 99.91 & 99.2& 99.62 & 99.53  & 99.91 & 94.7&95.7&71.3& 99.7&  99.2&99.6&99.9 & 99.81\\
Glue& 96.82 & 95.7& 99.61 & 98.94 & 99.61 & 91.9&87.2&37.5& 48.0&  88.9&98.1&99.3 & \bf 99.52 \\
Holep. & 86.87 & 81.9& 89.72 & 91.15 & 94.77 & 75.4&-&54.9&  69.7&  78.6& 94.3&96.5 & 91.91 \\
Iron & 100.0 & 98.9& 97.85 & 98.06 & 98.67 & 94.5&-&89.2& 97.4&  91.7&98.1&97.8 & \bf 98.47 \\
Lamp & 96.84 & 99.3& 97.79 & 99.14 & 99.14 & 96.6& -&87.6& 97.8&  94.0& 97.9&98.7 & \bf 99.71 \\
Phone & 94.69 & 92.4& 90.65 & 92.35 & 95.28 & 89.2&-&60.6& 76.0&  70.8& 96.4&97.3 & 93.86 \\ 
\hline
Mean & 95.15 & 86.3& 91.36 & 93.69 & 95.9 & 82.9& 70.73&63.6& 76.9&  86.0&96.6&97.8 & 95.41 \\ 
\hline
\end{tabular}}
\caption{Comparison with state-of-the-arts on the LM w.r.t. the metric of ADD(S)-0.1d. Noted that Gen6D$^{\dagger}$ uses a refinement strategy to train on a subset of LM, NeRF-Pose$^{\dagger}$ is trained on relative camera pose annotations. The best compared with CAD-free methods are in \textbf{bold}, the best compared with CAD-based methods are in {\bf\textit{italic bold}}.}
\label{tablelm}
\end{table*}

\begin{table}[t]
\centering
\resizebox{0.4\textwidth}{!}{
\begin{tabular}{c|cccc}
\hline
Object &OnePose& OnePose++&  GS-Pose&\textbf{Ours}\\ \hline
Ape &35.2& 97.3 &  97.5& \bf98.67\\
Bvise&94.4& \bf99.6&  98.5 &98.64\\
Camera &96.8& \bf99.6&  99.0& 99.31\\
Can &87.4& 99.2
&  97.6&\bf99.51\\
Cat &77.2& 98.7
&  99.0&\bf99.40\\
Driller &76.0&  93.1
&  91.9&\bf99.21\\
Duck &73.0& 97.7
&  97.6&\bf98.59\\
Eggbox&89.9& \bf98.7&  96.9&97.18\\
Glue&55.1& 51.8
&  96.8&\bf99.03\\
Holep. &79.1&  98.6 
&  98.2&\bf99.33\\
Iron &92.4 & \bf98.9&  96.8&98.06\\
Lamp &88.9& \bf98.8&  90.0&95.20\\
Phone &69.4& 94.5
&  91.1&\bf98.49\\ 
\hline
Mean &78.1 & 94.3&  96.2&\bf98.51\\ 
\hline
\end{tabular}
}
\caption{Comparison with state-of-the-arts on the LM w.r.t. the metric of Proj@5pix. Noted that all the other methods use YOLOv5~\cite{Ultralyticsyolov5} as the object detector and Ours uses YOLOv3~\cite{redmon2018yolov3}. We highlight the best in \textbf{bold}.}
\label{tablepj}
\end{table}

\begin{table*}[t]
\centering
\resizebox{0.97\textwidth}{!}{
\begin{tabular}{c|ccccc|cc|cccc}
\hline
Object & PoseCNN & PVNet & HybridPose& GDR-Net& SO-Pose & GDR-Net& SO-Pose &  RLLG&NeRF-Pose & NeRF-Pose$^{\dagger}$&  {\bf Ours} \\ \hline
CAD & \multicolumn{7}{c|}{\textit{w/} CAD} &\multicolumn{4}{c}{\textit{w/o} CAD} \\ \hline
Training & \multicolumn{5}{c|}{\textit{real + syn}} & \multicolumn{2}{c|}{\textit{real+pbr}} &\multicolumn{4}{c}{\textit{real+gen}} \\
\hline
Ape & 9.6 & 15.8 & 20.9 & 39.3 & 46.3 & 46.8 & 48.4 & 7.10&46.9 & 49.7& 35.13\\
Can & 45.2 &  63.3 & 75.3 & 79.2 &  81.1 & 90.8 & 85.8 &  40.6&86.2  & 86.4& 84.34 \\
Cat & 0.9 &  16.7 & 24.9 &  23.5 & 18.7 & 40.5 & 32.7 &  15.6&27.1 & 26.9& 23.34\\
Driller & 41.4 &  65.7 & 70.2 & 71.3 &  71.3 & 82.6 & 77.4 &  43.9&65.8 & 66.2& \bf84.68 \\
Duck & 19.6 &  25.2 & 27.9 & 44.4 & 43.9 & 46.9 & 48.9 & 12.9&29.9  & 36.9 & \bf43.48\\
Eggbox & 22.0 &  50.2 & 52.4 & 58.2 & 46.6 & 54.2 & 52.4 &  46.4&24.9  & 24.4& 44.68 \\
Glue & 38.5 &  49.6 & 53.8 & 49.3 &  63.3 & 75.8& 78.3 & 51.7&66.3 & 70.9& 69.77\\
Holep. & 22.1 & 36.1 & 54.2 &  58.7 &  62.9 & 60.1 & 75.3 & 24.5&46.4 & 49.8 & \bf 54.79 \\
\hline
Mean & 24.9 & 40.8 & 47.5 &  53.0 & 54.3 & 62.2 & 62.3 & 30.3&49.2 & 51.4& \bf55.03\\
\hline
\end{tabular}}
\caption{Comparison with state-of-the-arts on the LM-O w.r.t. the metric of ADD(S)-0.1d. \textquotedblleft real\textquotedblright\ is the real data provided by LM-O,  \textquotedblleft syn\textquotedblright\ is the blender synthetic data~\cite{li2018deepim}, \textquotedblleft pbr\textquotedblright\ is the physical-based rendering data~\cite{Denninger2023}, \textquotedblleft gen\textquotedblright\ is the model generated data. NeRF-Pose$^{\dagger}$ is trained on relative camera pose annotations. We highlight the best in \textbf{bold}.}
\label{tablelmo}
%\vspace{-0.5cm}
\end{table*}

\begin{table}[t]
\centering
\resizebox{0.35\textwidth}{!}{
\begin{tabular}{c|cc}
\hline
Object &Individual object& Occluded object\\ \hline
Ape &26.84& \bf35.13\\
Can &72.83& \bf84.34\\
Cat &14.57& \bf23.34\\
Driller &60.71&  \bf84.68\\
Duck &31.15& \bf43.48\\
Eggbox&19.32&\bf 44.68\\
Glue&38.43& \bf69.77\\
Holep. &41.4&  \bf54.79\\ 
\hline
Mean &38.15& \bf55.03\\ 
\hline
\end{tabular}
}
\caption{Comparison of individual object rendering and occluded object rending on the LM-O w.r.t. the ADD(S)-0.1d.}
\label{tableocc}
\end{table}

\section{Experiments}\label{exp}

In this section, extensive experiments are conducted to demonstrate the competitive performance of the proposed SGPose.  

\subsubsection{Datasets}

The proposed SGPose is evaluated on two commonly-used datasets, which are LM \cite{hinterstoisser2012model} and LM-O \cite{brachmann2014learning}.
LM is a standard benchmark for 6D object pose estimation of textureless objects, which consist of individual sequences of 13 objects 
%(i.e., ape, benchvise, camera, can, cat, driller, duck, eggbox, glue, holepuncher, iron, lamp and phone) 
of various sizes in the scenes with strong clutters and slight occlusion. Each contains about 1.2k real images with pose annotations. The dataset is split as $15\%$ for training and $85\%$ for testing.
LM-O is an extension of LM, from which to annotate one sequence of 8 objects with more severe occlusions of various degrees.
%(without benchvise, camera, iron, lamp and phone compared with LM).
For the LM dataset, approximately 1k  images with 2D-3D correspondence maps are rendered for each object.
For LM-O, which is designed for pose estimation in scenarios with occlusions, 50k images with significant occlusions on transparent backgrounds are rendered and involved in training with a ratio of 10:1, by following the convention of CAD-based methods for a fair comparison~\cite{wang2021gdr}.

\subsubsection{Implementation Details}

\begin{table}[t]
\centering
\resizebox{0.4\textwidth}{!}{
\begin{tabular}{c|c|c}
\hline
\# of real views& \# of synthetic views & Training result\\ \hline
30 & 0 & not converge\\
33 & 0 & converge\\
20 & 20 &  converge\\ 
10 & 20 &  converge \\
\hline
\end{tabular}
}
\caption{Ablation study on the effectiveness of synthetic view warping for the object \textbf{ape} in the LM.}
\label{tablewarp}
\end{table}

\begin{table*}[t]
\centering
\resizebox{0.9\textwidth}{!}{
\begin{tabular}{c|cc|cc|cc|cc|cc|cc|cc}
% \toprule [0.5mm]
\hline
 & \multicolumn{2}{c|}{benchvise} & \multicolumn{2}{c|}{camera} & \multicolumn{2}{c|}{cat} & \multicolumn{2}{c|}{can} & \multicolumn{2}{c|}{duck} & \multicolumn{2}{c|}{glue} & \multicolumn{2}{c}{Mean}\\
 \cline{2-15}
 & w/o. & w. & w/o. & w. & w/o. & w. & w/o. & w. & w/o. & w. & w/o. & w. & w/o. & w.\\
\hline
1mm   & \textbf{0.382} &0.377 & 0.503 & \textbf{0.553} & 0.417 & 0.417 & 0.383 
& \textbf{0.410} &0.379 & \textbf{0.398} &0.332 & \textbf{0.346} & 0.395 &\textbf{0.409}\\
3mm   & 0.570 & \textbf{0.584} & 0.710 & \textbf{0.776} & 0.656 & \textbf{0.663} & 0.597 & \textbf{0.644} &0.596 & \textbf{0.627} &0.545 & \textbf{0.564} & 0.615 
& \textbf{0.625}\\
5mm   & 0.831  & \textbf{0.857} & 0.921 & \textbf{0.951} & 0.936 & \textbf{0.948} & 0.911 
& \textbf{0.929} &0.960 & \textbf{0.978} &0.853 &\textbf{0.858} & 0.904 &0.904 \\
\hline
% \toprule [0.5mm]
\end{tabular}}
\caption{Ablations of online pruning with selected objects on LM dataset.}
\label{table_bop}
\end{table*}

\begin{table}[t]
\centering
\resizebox{0.45\textwidth}{!}{
\begin{tabular}{c|cc|cc|cc}
% \toprule [0.5mm]
\hline
 & \multicolumn{2}{c|}{can} & \multicolumn{2}{c|}{cat}  & \multicolumn{2}{c}{glue} \\
 \cline{2-7}
 & w/o. & w. & w/o. & w. & w/o. & w. \\
\hline
ADD(S)-0.1d   & 96.85 & \bf97.34 & 51.90 & \bf88.12 & \bf99.13& 99.03 \\
Proj@5pix   & 98.33 & \bf98.43 & 46.91 & \bf99.10 & 93.63 & \bf96.91 \\
\hline
% \toprule [0.5mm]
\end{tabular}}
\caption{Pose estimation comparison of selected objects on LM dataset w.r.t. ADD(S)-0.1d and Proj@5pix without and with online pruning.}
\label{tab_pose}
\end{table}

Ten real images and corresponding pose annotations are taken as input of the proposed geometric-aware object Gaussian, which are selected from real data in LM.
Different selection strategies are conducted, e.g., selecting samples uniformly, randomly, in term of maximum rotation differences and maximum Intersection over Union (IoU). 
The simple uniform selection is adopted in our method taking account of real-world practice.

The proposed object Gaussian tailors respective optimization strategies to the supervision signals.
The number of points of random initialization is set as 4,096.
The geometric-consistent loss involves in training at iteration 3,000 and the normal loss is enabled at iteration 7,000 as in~\cite{huang20242d}.
Ten sparse views are given, and two synthetic views are created online around each give view at iteration 4,999, that is, 30 images involve in training for each object.
The synthetic image warping spans 40\% of training phase once it is activated, which updates the model alternatively with the image rendering loss.
Both image rendering loss and image warping loss are endowed with equal weights, dominating the training process.

Noted that it’s a little tricky to conduct pruning effectively in our problem settings.
Regions of Interests (RoIs) are remained for training and backgrounds are masked out, it is possible that pruning techniques working well for distant floaters remove part of the foreground objects mistakenly, which is more pronounced when the objects are thin and tall.
Thus, the objects phone and driller do not apply pruning empirically.

For the occluded scene generation for LM-O dataset, we utilize pose annotations from PBR (Physically Based Rendering)~\cite{Denninger2023} for image rendering.
Each object is rendered onto the image using its respective geometric-aware object Gaussian, with all objects rendered sequentially in a single image. 
Realizing that individual object Gaussian do not inherently represent occlusions, we overlay the rendered images with the visible masks from PBR data to simulate occlusion scenarios effectively. 

The rendered masks, crafted from the proposed geometric-aware object Gaussian, is generated by mapping color images to a binary format. 
That is, assigning \textquotedblleft 1\textquotedblright\ to pixels within the object and \textquotedblleft 0\textquotedblright\ to those outside, thus producing a boolean array congruent with the original image's dimensions. 
It is possible that incorporating an extra mask loss for supervision could improve the performance of mask rendering in future work.

The 2D bounding boxes for pose estiamtion are obtained by borrow an off-the-shelf object detector yolov3~\cite{redmon2018yolov3}.
%for LM dataset, the proposed SGPose is trained on training set of real data, by strictly following the training/testing split as in~\cite{wang2021gdr}; for LM-O dataset, our method is trained on all real data of LM by following previous works for a fair comparison.

\subsubsection{Evaluation Metrics}
We evaluate our method with the most commonly used metrics including ADD(S)-0.1d and Proj@5pix.
%$n^{\circ} n\;cm$ \cite{shotton2013scene}.
ADD(S)-0.1d measures the mean distance between the model points transformed from the estimated pose and the ground truth. If the percentage of mean distance lies below 10\% of the object’s diameter (0.1d), the estimated pose is regarded as correct. For symmetric objects with pose ambiguity, ADD(-S) measures the deviation to the closet model point \cite{hinterstoisser2012model, hodan2020epos}. 
Proj@5pix computes the mean distance between the projection of 3D model points with given predicted and ground truth object poses. The estimated pose is considered correct if the mean projection distance is less than 5 pixels.

%Additionally, $n^{\circ}\;n\;cm$ measures the percentage of estimated poses with rotation error less than $n^{\circ}$ and translation error below $n\;cm$. 
% For YCB-V, the Area Under Curve (AUC) of ADD(-S) is computed by varying the distance threshold under $10cm$ \cite{xiang2018posecnn}.

\subsection{Comparison with State-of-the-Arts}

\subsubsection{Results on LM}

\begin{figure*}[htbp]
\centering
\includegraphics[width=0.99\textwidth]{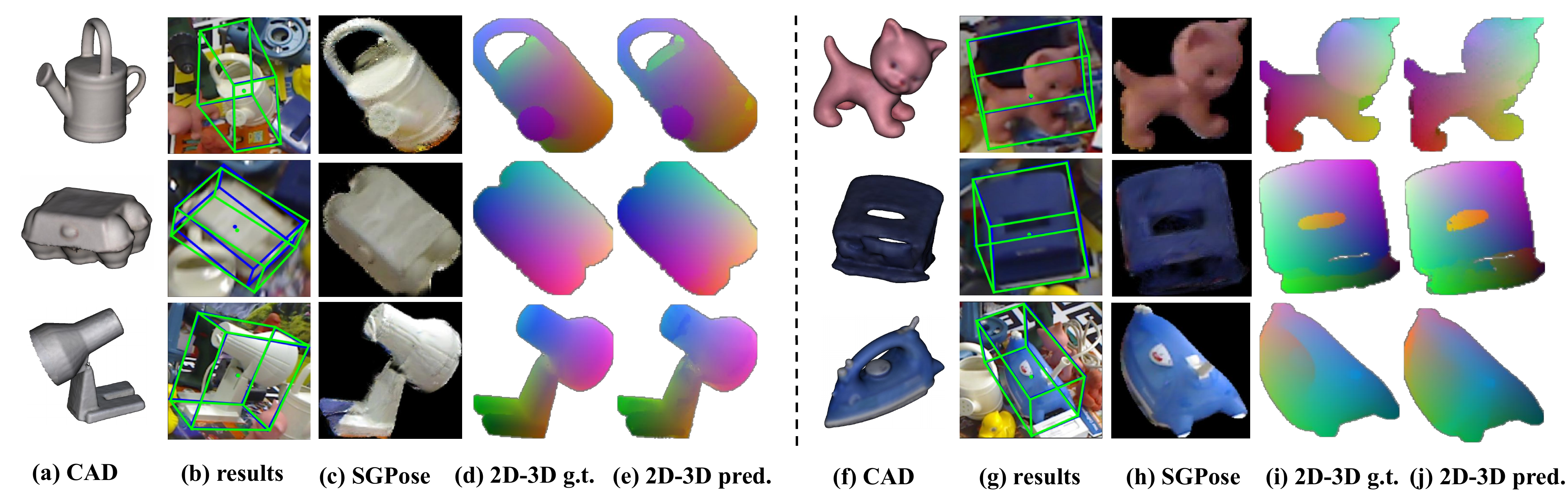}
\caption{Qualitative results on LM. Column (a) and (f) show the CAD models. Column (b) and (g) illustrate the predicted object poses (green) and ground truth poses (blue). Column (c) and (h) are the rendered images from our object Gaussian. Column (d) and (i) are the generated 2D-3D correspondence maps from our object Gaussian, which are also \textit{g.t. }of regression network. Column (e) and (j) are the predicted 2D-3D correspondence maps from our regression network. }
\label{fig2}
\end{figure*}

The proposed method is compared with the CAD-based methods DPOD~\cite{zakharov2019dpod}, PVNet~\cite{peng2019pvnet}, CDPN~\cite{li2019cdpn}, GDR-Net~\cite{wang2021gdr}, SO-Pose~\cite{di2021so} and CAD-free methods RLLG~\cite{cai2020reconstruct} , Gen6D~\cite{liu2022gen6d}, OnePose~\cite{sun2022onepose}, OnePose++~\cite{he2022onepose++}, GS-Pose~\cite{cai2024gs}, and Nerf-Pose~\cite{li2023nerf} on metric of ADD(S)-0.1d and Proj@5pix.
As shown in Tab.~\ref{tablelm}, even under the setting of sparse view training, our method achieves comparable performance compared to most CAD-free baselines that are trained with more than 100 views, and is on par with the CAD-based methods. 
Notably, our proposed object Gaussian is trained on only 10 views, whereas Nerf-Pose uses 156 views for OBJ-NeRF training.  
In brief, the objects where our method outperforms the best CAD-free method (i.e., NeRF-Pose$^{\dagger}$) are highlighted in bold, and where it surpasses the best CAD-based method (i.e., SO-Pose) are in italic bold. 
Noteworthy, Gen6D$^{\dagger}$ is refined on a subset of the LM dataset, and NeRF-Pose$^{\dagger}$ is trained on relative camera pose annotations.
As show in Tab.~\ref{tablepj}, SGPose demonstrates an impressive 98.51\% average performance using only 10 given views, outperforming all baselines according to the metric of Proj@5pix. Noted that our method uses YOLOv3~\cite{redmon2018yolov3} as the object detector, while all the others use the more recent YOLOv5~\cite{Ultralyticsyolov5}.
%state-of-the-art methods in terms of ADD(-S), $2^{\circ}\;2cm$, and $5^{\circ}\;5cm$.
%Under ADD(-S) 0.05d, 0.1d, and $5^{\circ}5 cm$, our method outperforms all the baseline methods. 
%Notably, ADD(-S) 0.02d is increased from $35.5$ to $44.4$, with a substantial enhancement up to $25\%$.
%
\subsubsection{Results on LM-O}

As demonstrated in Tab.~\ref{tablelmo}, our method is compared with state-of-the-arts w.r.t. the metric of average recall (\%) of ADD(-S). 
Among CAD-based methods, \textquotedblleft real+pbr\textquotedblright\ outperforms \textquotedblleft real+syn\textquotedblright\ because \textquotedblleft pbr\textquotedblright\ data~\cite{Denninger2023} incorporate occlusions in object placement, with random textures, materials, and lighting, simulating a more natural environment compared to individually rendered synthetic data. Given the heavy occlusions typical of the LM-O dataset, training with \textquotedblleft pbr\textquotedblright\ data significantly enhances performance. 
In our setting, we do not have access to CAD models nor do we leverage \textquotedblleft pbr\textquotedblright\ data. 
Instead, we render the synthetic images that replicates the occlusion scenarios found in the LM-O dataset, using our proposed object Gaussian.
Exemplary, we exceed Nerf-Pose~\cite{li2023nerf} by 5.83\% with 55.03\% compared to 49.2\%, also rival NeRF-Pose$^{\dagger}$, which is trained on relative camera poses instead of ground truth pose annotations, by 3.63\%. Impressively, we even slightly outperform the best CAD-based method SO-Pose~\cite{di2021so}.

\subsection{Ablations}

\subsubsection{Qualitative comparison of 2D-3D Correspondence}

The qualitative results of selected objects are presented in Fig. \ref{fig2}, where the transformed 3D bounding boxes are overlaid with the corresponding images. 
As observed, the predicted poses (green boxes) mostly align with the ground truth (blue boxes). 
The images are cropped and zoomed into the area of interest for better visualization.
The rendered images are exhibited in column (c) and (h), compared to the reference CAD models in column (a), our object Gaussian successfully retains both the silhouette and the details of the objects. 
The accurate geometric shapes rendered from the object Gaussian ensure the performance of the subsequent pose estimation.  
Nonetheless, given the inherently challenging nature of sparse view reconstruction, some imperfections in the predicted shapes are also evident. 

\subsubsection{Training with Occlusions}

Since LM-O is a more challenging dataset presenting complex occlusions of objects, the integration of synthetic data that captures diverse poses and realistic occlusions is beneficial for enhancing performance. 
We thus avail the object Gaussian to render such images to enrich training.
Quantitatively, as shown in Tab.~\ref{tableocc},  two different synthetic data are rendered for training, "Occluded object" indicates images containing multiple objects with occlusions, whereas "Individual object" signifies images with a single unoccluded object. 
We observe that the use of "Occluded object" rendering improves the performance of the proposed SGPose by large margins under all the objects.
%The qualitative results are demystified in the supplementary material.

\begin{figure*}[htbp]
\centering
\includegraphics[width=0.85\textwidth]{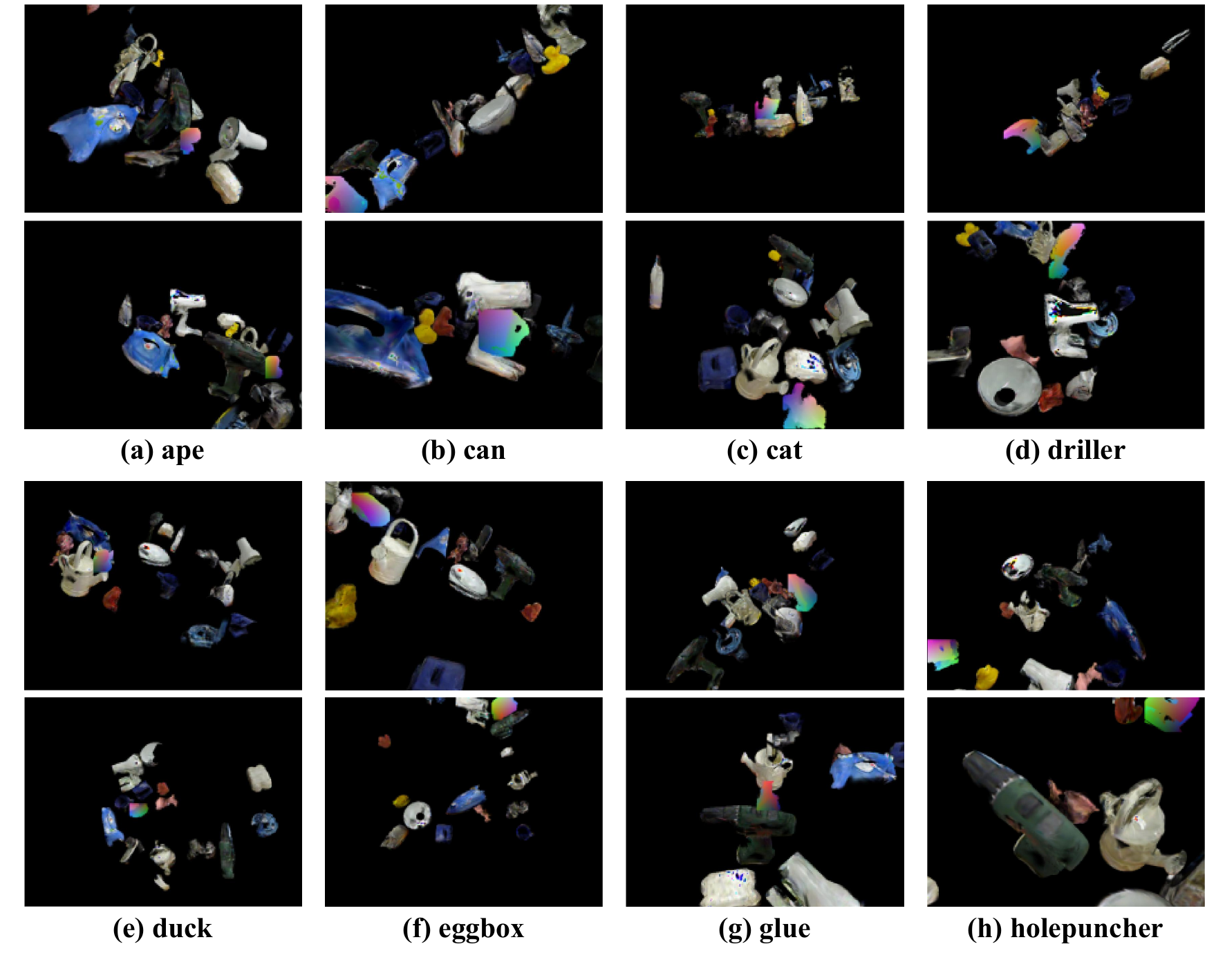} % Reduce the figure size so that it is slightly narrower than the column.
\caption{Qualitative results for each object of synthetic data for the LM-O dataset, where the 2D-3D correspondences are projected onto the target object for visualization.}
\label{fig_occ}
\end{figure*}

For the LM-O dataset, the proposed SGPose generates images and 2D-3D correspondences that demonstrate a diverse range of poses and realistic occlusions, as shown in Fig.~\ref{fig_occ}. 
In the training process, the synthetic images are integrated at a 10:1 ratio with real images, meaning that for every ten real images, one synthetic image is included.
The 2D-3D correspondence maps are projected onto the target object in the images for visualization. 
Compared to training with individual object rendering, the inclusion of occluded object rendering remarkably enhances the model's performance in complex scenarios where objects have partial visibility.

\subsubsection{Effectiveness of Synthetic View Warping}

As shown in Tab.~\ref{tablewarp}, successful reconstruction of the object ape in LM requires a minimum of 33 images without synthetic view warping. 
By synthesising 20 novel view alone with 10 given images, the proposed SGPose maintains the performance. 
This demonstrates the effectiveness of synthetic view warping in reducing the reliance on real images.

\subsubsection{Effectiveness of Online Pruning}

Point cloud accuracy is quantified as the proportion of reconstructed point clouds that fall within a specified distance threshold (e.g., \textit{3mm}) relative to the ground truth point clouds, where the vertices of the object meshes serve as the ground truth reference~\cite{sarlin2023pixel, schops2017multi}.
The point cloud accuracy is evaluated without and with online pruning for our proposed object Gaussian, following the established protocols in \cite{sarlin2023pixel} and \cite{he2022onepose++}.  
The results presented in Tab.~\ref{table_bop} demonstrate that the online pruning removes outliers from sparse view object Gaussian reconstruction, resulting in a more accurate and compact representation.
Additionally, pose estimation comparison of selected objects w.r.t. ADD(S)-0.1d and Proj@5pix without and with online pruning is presented in Tab.~\ref{tab_pose}.
Besides, sparse view reconstruction poses a challenge for object with thin and long geometry, such as lamp and glue. The timely application of online pruning, initiated as divergence threatens, ensures the model to be reconstructed successfully.

\subsection{Qualitative Results of Synthetic View Warping}

\begin{figure*}[htbp]
\centering
\includegraphics[width=0.7\textwidth]{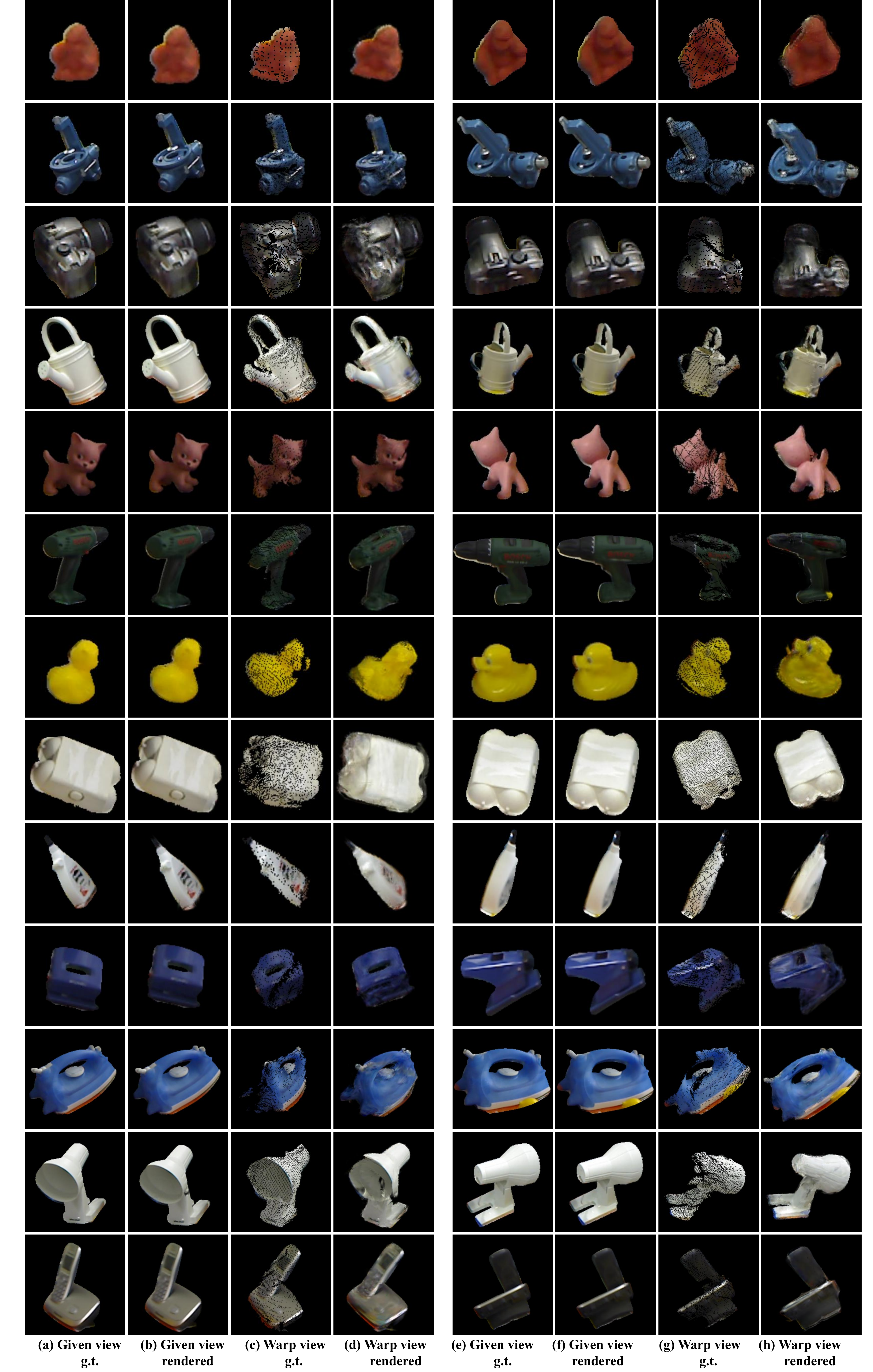} % Reduce the figure size so that it is slightly narrower than the column.
\caption{Qualitative results for selected synthetic views. Column (a) and (e) display ground truth of given views; (b) and (f) show SGPose rendered images of given views. Synthetic view ground truths are in (c) and (g), with their corresponding rendered images in (d) and (h). Synthetic views are generated by applying rotation perturbations of up to ±15° and translation perturbations of ±0.01 m along the x and y axes, and ±0.05 m along the z-axis to the given views.}
\label{fig_warp}
\end{figure*}

The synthetic view for each real image is generated by introducing a controlled amount of noise to the given view, ensuring that the synthetic images retain a realistic and plausible appearance. The perturbation parameters are carefully chosen to keep the object within the camera's field of view.
Specifically, the Euler angles for rotation perturbation are sampled from a normal distribution with a standard deviation of 15°, capped at an upper limit of 45°. The translation perturbation along each axis is independently sampled from a normal distribution with standard deviations of 0.01 m for the x and y axes, and 0.05 m for the z-axis, respectively.
Fig.~\ref{fig_warp} displays the qualitative results. Column (a) and (e) presents the ground truth of given views, while column (b) and (f) shows the corresponding rendered images from SGPose. Columns (c) and (g) illustrate the ground truth of synthetic views, and columns (d) and (h) exhibit the rendered synthetic images, respectively. 
The rendered results are obtained at the iteration 30k upon completion of the training.

\FloatBarrier

\FloatBarrier

\subsection{Implementation and Runtime Analysis}

The experiments are conduct on an platform with an Intel(R) Xeon(R) Gold 5220R 2.20GHz CPU and Nvidia RTX3090 GPUs of 24GB Memory.
Given ten $640\times480$ images as input, the object Gaussian costs about 10 minutes to reconstruct one object and real-time renders image, mask, and 2D-3D correspondence.
An ImageNet~\cite{deng2009imagenet} pre-trained ConvNeXt~\cite{liu2022convnet} network is leveraged as the backbone of our pose regression network, for a $640\times480$ image, the proposed SGPose takes about 24ms for inference.

\section{Limitations}

In future work, we plan to reduce the training time to enable portable online reconstruction and pose estimation, thereby facilitating real-time, end-to-end pose estimation suitable for real-world applications.

\section{Conclusion}

The proposed SGPose presents a monocular object pose estimation framework, effectively addressing the limitations of traditional methods that rely on CAD models. 
By introducing a novel approach that requires as few as ten reference views, the derived geometric-aware depth guides the object-centric Gaussian model to perform synthetic view warping and online pruning effectively, showcasing its robustness and applicability in real-world scenarios under sparse view constrains. 
The occlusion data rendered from the proposed object Gaussian substantially enhances the performance of post estimation, set our SGPose a state-of-the-art on the Occlusion LM-O dataset.

\appendix

\bigskip

\bibliography{aaai25}

\begin{thebibliography}{58}
\providecommand{\natexlab}[1]{#1}

\bibitem[{Ahmadyan et~al.(2021)Ahmadyan, Zhang, Ablavatski, Wei, and
  Grundmann}]{ahmadyan2021objectron}
Ahmadyan, A.; Zhang, L.; Ablavatski, A.; Wei, J.; and Grundmann, M. 2021.
\newblock Objectron: A large scale dataset of object-centric videos in the wild
  with pose annotations.
\newblock In \emph{Proceedings of the IEEE/CVF conference on computer vision
  and pattern recognition}, 7822--7831.

\bibitem[{Azad, Asfour, and Dillmann(2007)}]{azad2007stereo}
Azad, P.; Asfour, T.; and Dillmann, R. 2007.
\newblock Stereo-based 6d object localization for grasping with humanoid robot
  systems.
\newblock In \emph{2007 IEEE/RSJ International Conference on Intelligent Robots
  and Systems}, 919--924. IEEE.

\bibitem[{Botsch et~al.(2005)Botsch, Hornung, Zwicker, and
  Kobbelt}]{botsch2005high}
Botsch, M.; Hornung, A.; Zwicker, M.; and Kobbelt, L. 2005.
\newblock High-quality surface splatting on today's GPUs.
\newblock In \emph{Proceedings Eurographics/IEEE VGTC Symposium Point-Based
  Graphics, 2005.}, 17--141. IEEE.

\bibitem[{Brachmann et~al.(2014)Brachmann, Krull, Michel, Gumhold, Shotton, and
  Rother}]{brachmann2014learning}
Brachmann, E.; Krull, A.; Michel, F.; Gumhold, S.; Shotton, J.; and Rother, C.
  2014.
\newblock Learning 6d object pose estimation using 3d object coordinates.
\newblock In \emph{European conference on computer vision}, 536--551. Springer.

\bibitem[{Cai, Heikkil{\"a}, and Rahtu(2024)}]{cai2024gs}
Cai, D.; Heikkil{\"a}, J.; and Rahtu, E. 2024.
\newblock Gs-pose: Cascaded framework for generalizable segmentation-based 6d
  object pose estimation.
\newblock \emph{arXiv preprint arXiv:2403.10683}.

\bibitem[{Cai and Reid(2020)}]{cai2020reconstruct}
Cai, M.; and Reid, I. 2020.
\newblock Reconstruct locally, localize globally: A model free method for
  object pose estimation.
\newblock In \emph{Proceedings of the IEEE/CVF Conference on Computer Vision
  and Pattern Recognition}, 3153--3163.

\bibitem[{Chen et~al.(2020)Chen, Dong, Song, Geiger, and
  Hilliges}]{chen2020category}
Chen, X.; Dong, Z.; Song, J.; Geiger, A.; and Hilliges, O. 2020.
\newblock Category level object pose estimation via neural
  analysis-by-synthesis.
\newblock In \emph{Computer Vision--ECCV 2020: 16th European Conference,
  Glasgow, UK, August 23--28, 2020, Proceedings, Part XXVI 16}, 139--156.
  Springer.

\bibitem[{Deng et~al.(2009)Deng, Dong, Socher, Li, Li, and
  Fei-Fei}]{deng2009imagenet}
Deng, J.; Dong, W.; Socher, R.; Li, L.-J.; Li, K.; and Fei-Fei, L. 2009.
\newblock Imagenet: A large-scale hierarchical image database.
\newblock In \emph{2009 IEEE conference on computer vision and pattern
  recognition}, 248--255. Ieee.

\bibitem[{Denninger et~al.(2023)Denninger, Winkelbauer, Sundermeyer, Boerdijk,
  Knauer, Strobl, Humt, and Triebel}]{Denninger2023}
Denninger, M.; Winkelbauer, D.; Sundermeyer, M.; Boerdijk, W.; Knauer, M.;
  Strobl, K.~H.; Humt, M.; and Triebel, R. 2023.
\newblock BlenderProc2: A Procedural Pipeline for Photorealistic Rendering.
\newblock \emph{Journal of Open Source Software}, 8(82): 4901.

\bibitem[{Di et~al.(2021)Di, Manhardt, Wang, Ji, Navab, and Tombari}]{di2021so}
Di, Y.; Manhardt, F.; Wang, G.; Ji, X.; Navab, N.; and Tombari, F. 2021.
\newblock So-pose: Exploiting self-occlusion for direct 6d pose estimation.
\newblock In \emph{Proceedings of the IEEE/CVF International Conference on
  Computer Vision}, 12396--12405.

\bibitem[{Fridovich-Keil et~al.(2022)Fridovich-Keil, Yu, Tancik, Chen, Recht,
  and Kanazawa}]{fridovich2022plenoxels}
Fridovich-Keil, S.; Yu, A.; Tancik, M.; Chen, Q.; Recht, B.; and Kanazawa, A.
  2022.
\newblock Plenoxels: Radiance fields without neural networks.
\newblock In \emph{Proceedings of the IEEE/CVF conference on computer vision
  and pattern recognition}, 5501--5510.

\bibitem[{He et~al.(2022)He, Sun, Wang, Huang, Bao, and Zhou}]{he2022onepose++}
He, X.; Sun, J.; Wang, Y.; Huang, D.; Bao, H.; and Zhou, X. 2022.
\newblock Onepose++: Keypoint-free one-shot object pose estimation without CAD
  models.
\newblock \emph{Advances in Neural Information Processing Systems}, 35:
  35103--35115.

\bibitem[{Hinterstoisser et~al.(2012)Hinterstoisser, Lepetit, Ilic, Holzer,
  Bradski, Konolige, and Navab}]{hinterstoisser2012model}
Hinterstoisser, S.; Lepetit, V.; Ilic, S.; Holzer, S.; Bradski, G.; Konolige,
  K.; and Navab, N. 2012.
\newblock Model based training, detection and pose estimation of texture-less
  3d objects in heavily cluttered scenes.
\newblock In \emph{Asian conference on computer vision}, 548--562. Springer.

\bibitem[{Hodan, Barath, and Matas(2020)}]{hodan2020epos}
Hodan, T.; Barath, D.; and Matas, J. 2020.
\newblock Epos: Estimating 6d pose of objects with symmetries.
\newblock In \emph{Proceedings of the IEEE/CVF conference on computer vision
  and pattern recognition}, 11703--11712.

\bibitem[{Huang et~al.(2024)Huang, Yu, Chen, Geiger, and Gao}]{huang20242d}
Huang, B.; Yu, Z.; Chen, A.; Geiger, A.; and Gao, S. 2024.
\newblock 2d gaussian splatting for geometrically accurate radiance fields.
\newblock In \emph{ACM SIGGRAPH 2024 Conference Papers}, 1--11.

\bibitem[{Jain, Tancik, and Abbeel(2021)}]{jain2021putting}
Jain, A.; Tancik, M.; and Abbeel, P. 2021.
\newblock Putting nerf on a diet: Semantically consistent few-shot view
  synthesis.
\newblock In \emph{Proceedings of the IEEE/CVF International Conference on
  Computer Vision}, 5885--5894.

\bibitem[{Kehl et~al.(2017)Kehl, Manhardt, Tombari, Ilic, and
  Navab}]{kehl2017ssd}
Kehl, W.; Manhardt, F.; Tombari, F.; Ilic, S.; and Navab, N. 2017.
\newblock Ssd-6d: Making rgb-based 3d detection and 6d pose estimation great
  again.
\newblock In \emph{Proceedings of the IEEE international conference on computer
  vision}, 1521--1529.

\bibitem[{Kerbl et~al.(2023)Kerbl, Kopanas, Leimk{\"u}hler, and
  Drettakis}]{kerbl20233d}
Kerbl, B.; Kopanas, G.; Leimk{\"u}hler, T.; and Drettakis, G. 2023.
\newblock 3D Gaussian Splatting for Real-Time Radiance Field Rendering.
\newblock \emph{ACM Trans. Graph.}, 42(4): 139--1.

\bibitem[{Kopanas et~al.(2021)Kopanas, Philip, Leimk{\"u}hler, and
  Drettakis}]{kopanas2021point}
Kopanas, G.; Philip, J.; Leimk{\"u}hler, T.; and Drettakis, G. 2021.
\newblock Point-Based Neural Rendering with Per-View Optimization.
\newblock In \emph{Computer Graphics Forum}, volume~40, 29--43. Wiley Online
  Library.

\bibitem[{Labb{\'e} et~al.(2020)Labb{\'e}, Carpentier, Aubry, and
  Sivic}]{labbe2020cosypose}
Labb{\'e}, Y.; Carpentier, J.; Aubry, M.; and Sivic, J. 2020.
\newblock Cosypose: Consistent multi-view multi-object 6d pose estimation.
\newblock In \emph{Computer Vision--ECCV 2020: 16th European Conference,
  Glasgow, UK, August 23--28, 2020, Proceedings, Part XVII 16}, 574--591.
  Springer.

\bibitem[{Lee et~al.(2021)Lee, Lee, Kim, and Kweon}]{lee2021category}
Lee, T.; Lee, B.-U.; Kim, M.; and Kweon, I.~S. 2021.
\newblock Category-level metric scale object shape and pose estimation.
\newblock \emph{IEEE Robotics and Automation Letters}, 6(4): 8575--8582.

\bibitem[{Lepetit, Moreno-Noguer, and Fua(2009)}]{lepetit2009epnp}
Lepetit, V.; Moreno-Noguer, F.; and Fua, P. 2009.
\newblock Epnp: An accurate o (n) solution to the pnp problem.
\newblock \emph{International journal of computer vision}, 81(2): 155--166.

\bibitem[{Li et~al.(2023)Li, Vutukur, Yu, Shugurov, Busam, Yang, and
  Ilic}]{li2023nerf}
Li, F.; Vutukur, S.~R.; Yu, H.; Shugurov, I.; Busam, B.; Yang, S.; and Ilic, S.
  2023.
\newblock Nerf-pose: A first-reconstruct-then-regress approach for
  weakly-supervised 6d object pose estimation.
\newblock In \emph{Proceedings of the IEEE/CVF International Conference on
  Computer Vision}, 2123--2133.

\bibitem[{Li et~al.(2018)Li, Wang, Ji, Xiang, and Fox}]{li2018deepim}
Li, Y.; Wang, G.; Ji, X.; Xiang, Y.; and Fox, D. 2018.
\newblock Deepim: Deep iterative matching for 6d pose estimation.
\newblock In \emph{Proceedings of the European Conference on Computer Vision
  (ECCV)}, 683--698.

\bibitem[{Li, Wang, and Ji(2019)}]{li2019cdpn}
Li, Z.; Wang, G.; and Ji, X. 2019.
\newblock Cdpn: Coordinates-based disentangled pose network for real-time
  rgb-based 6-dof object pose estimation.
\newblock In \emph{Proceedings of the IEEE/CVF International Conference on
  Computer Vision}, 7678--7687.

\bibitem[{Liu et~al.(2022{\natexlab{a}})Liu, Zhang, Zhang, Fu, Tang, Liang,
  Tang, Cheng, Zhang, Wang, and Ji}]{liu2022gdrnpp_bop}
Liu, X.; Zhang, R.; Zhang, C.; Fu, B.; Tang, J.; Liang, X.; Tang, J.; Cheng,
  X.; Zhang, Y.; Wang, G.; and Ji, X. 2022{\natexlab{a}}.
\newblock GDRNPP.
\newblock \url{https://github.com/shanice-l/gdrnpp_bop2022}.

\bibitem[{Liu et~al.(2022{\natexlab{b}})Liu, Wen, Peng, Lin, Long, Komura, and
  Wang}]{liu2022gen6d}
Liu, Y.; Wen, Y.; Peng, S.; Lin, C.; Long, X.; Komura, T.; and Wang, W.
  2022{\natexlab{b}}.
\newblock Gen6d: Generalizable model-free 6-dof object pose estimation from rgb
  images.
\newblock In \emph{European Conference on Computer Vision}, 298--315. Springer.

\bibitem[{Liu et~al.(2022{\natexlab{c}})Liu, Mao, Wu, Feichtenhofer, Darrell,
  and Xie}]{liu2022convnet}
Liu, Z.; Mao, H.; Wu, C.-Y.; Feichtenhofer, C.; Darrell, T.; and Xie, S.
  2022{\natexlab{c}}.
\newblock A convnet for the 2020s.
\newblock In \emph{Proceedings of the IEEE/CVF conference on computer vision
  and pattern recognition}, 11976--11986.

\bibitem[{Manhardt, Kehl, and Gaidon(2019)}]{manhardt2019roi}
Manhardt, F.; Kehl, W.; and Gaidon, A. 2019.
\newblock Roi-10d: Monocular lifting of 2d detection to 6d pose and metric
  shape.
\newblock In \emph{Proceedings of the IEEE/CVF Conference on Computer Vision
  and Pattern Recognition}, 2069--2078.

\bibitem[{Mildenhall et~al.(2021)Mildenhall, Srinivasan, Tancik, Barron,
  Ramamoorthi, and Ng}]{mildenhall2021nerf}
Mildenhall, B.; Srinivasan, P.~P.; Tancik, M.; Barron, J.~T.; Ramamoorthi, R.;
  and Ng, R. 2021.
\newblock Nerf: Representing scenes as neural radiance fields for view
  synthesis.
\newblock \emph{Communications of the ACM}, 65(1): 99--106.

\bibitem[{Niemeyer et~al.(2022)Niemeyer, Barron, Mildenhall, Sajjadi, Geiger,
  and Radwan}]{niemeyer2022regnerf}
Niemeyer, M.; Barron, J.~T.; Mildenhall, B.; Sajjadi, M.~S.; Geiger, A.; and
  Radwan, N. 2022.
\newblock Regnerf: Regularizing neural radiance fields for view synthesis from
  sparse inputs.
\newblock In \emph{Proceedings of the IEEE/CVF Conference on Computer Vision
  and Pattern Recognition}, 5480--5490.

\bibitem[{Oberweger, Rad, and Lepetit(2018)}]{oberweger2018making}
Oberweger, M.; Rad, M.; and Lepetit, V. 2018.
\newblock Making deep heatmaps robust to partial occlusions for 3d object pose
  estimation.
\newblock In \emph{Proceedings of the European conference on computer vision
  (ECCV)}, 119--134.

\bibitem[{Park et~al.(2020)Park, Mousavian, Xiang, and
  Fox}]{park2020latentfusion}
Park, K.; Mousavian, A.; Xiang, Y.; and Fox, D. 2020.
\newblock Latentfusion: End-to-end differentiable reconstruction and rendering
  for unseen object pose estimation.
\newblock In \emph{Proceedings of the IEEE/CVF conference on computer vision
  and pattern recognition}, 10710--10719.

\bibitem[{Park, Patten, and Vincze(2019)}]{park2019pix2pose}
Park, K.; Patten, T.; and Vincze, M. 2019.
\newblock Pix2pose: Pixel-wise coordinate regression of objects for 6d pose
  estimation.
\newblock In \emph{Proceedings of the IEEE/CVF international conference on
  computer vision}, 7668--7677.

\bibitem[{Pavlakos et~al.(2017)Pavlakos, Zhou, Chan, Derpanis, and
  Daniilidis}]{pavlakos20176}
Pavlakos, G.; Zhou, X.; Chan, A.; Derpanis, K.~G.; and Daniilidis, K. 2017.
\newblock 6-dof object pose from semantic keypoints.
\newblock In \emph{2017 IEEE international conference on robotics and
  automation (ICRA)}, 2011--2018. IEEE.

\bibitem[{Peng et~al.(2019)Peng, Liu, Huang, Zhou, and Bao}]{peng2019pvnet}
Peng, S.; Liu, Y.; Huang, Q.; Zhou, X.; and Bao, H. 2019.
\newblock PVNet: Pixel-wise voting network for 6DoF object pose estimation.
\newblock \emph{IEEE Transactions on Pattern Analysis and Machine
  Intelligence}, 14(8).

\bibitem[{Qi et~al.(2018)Qi, Liu, Wu, Su, and Guibas}]{qi2018frustum}
Qi, C.~R.; Liu, W.; Wu, C.; Su, H.; and Guibas, L.~J. 2018.
\newblock Frustum pointnets for 3d object detection from rgb-d data.
\newblock In \emph{Proceedings of the IEEE conference on computer vision and
  pattern recognition}, 918--927.

\bibitem[{Redmon and Farhadi(2018)}]{redmon2018yolov3}
Redmon, J.; and Farhadi, A. 2018.
\newblock Yolov3: An incremental improvement.
\newblock \emph{arXiv preprint arXiv:1804.02767}.

\bibitem[{Sarlin et~al.(2023)Sarlin, Lindenberger, Larsson, and
  Pollefeys}]{sarlin2023pixel}
Sarlin, P.-E.; Lindenberger, P.; Larsson, V.; and Pollefeys, M. 2023.
\newblock Pixel-perfect structure-from-motion with featuremetric refinement.
\newblock \emph{IEEE Transactions on Pattern Analysis and Machine
  Intelligence}.

\bibitem[{Sch\"{o}nberger and Frahm(2016)}]{schoenberger2016sfm}
Sch\"{o}nberger, J.~L.; and Frahm, J.-M. 2016.
\newblock Structure-from-Motion Revisited.
\newblock In \emph{Conference on Computer Vision and Pattern Recognition
  (CVPR)}.

\bibitem[{Sch\"{o}nberger et~al.(2016)Sch\"{o}nberger, Zheng, Pollefeys, and
  Frahm}]{schoenberger2016mvs}
Sch\"{o}nberger, J.~L.; Zheng, E.; Pollefeys, M.; and Frahm, J.-M. 2016.
\newblock Pixelwise View Selection for Unstructured Multi-View Stereo.
\newblock In \emph{European Conference on Computer Vision (ECCV)}.

\bibitem[{Schops et~al.(2017)Schops, Schonberger, Galliani, Sattler, Schindler,
  Pollefeys, and Geiger}]{schops2017multi}
Schops, T.; Schonberger, J.~L.; Galliani, S.; Sattler, T.; Schindler, K.;
  Pollefeys, M.; and Geiger, A. 2017.
\newblock A multi-view stereo benchmark with high-resolution images and
  multi-camera videos.
\newblock In \emph{Proceedings of the IEEE conference on computer vision and
  pattern recognition}, 3260--3269.

\bibitem[{Sigg et~al.(2006)Sigg, Weyrich, Botsch, and Gross}]{sigg2006gpu}
Sigg, C.; Weyrich, T.; Botsch, M.; and Gross, M.~H. 2006.
\newblock GPU-based ray-casting of quadratic surfaces.
\newblock In \emph{PBG@ SIGGRAPH}, 59--65.

\bibitem[{Sun et~al.(2022)Sun, Wang, Zhang, He, Zhao, Zhang, and
  Zhou}]{sun2022onepose}
Sun, J.; Wang, Z.; Zhang, S.; He, X.; Zhao, H.; Zhang, G.; and Zhou, X. 2022.
\newblock Onepose: One-shot object pose estimation without cad models.
\newblock In \emph{Proceedings of the IEEE/CVF Conference on Computer Vision
  and Pattern Recognition}, 6825--6834.

\bibitem[{Takikawa et~al.(2022)Takikawa, Evans, Tremblay, M{\"u}ller, McGuire,
  Jacobson, and Fidler}]{takikawa2022variable}
Takikawa, T.; Evans, A.; Tremblay, J.; M{\"u}ller, T.; McGuire, M.; Jacobson,
  A.; and Fidler, S. 2022.
\newblock Variable bitrate neural fields.
\newblock In \emph{ACM SIGGRAPH 2022 Conference Proceedings}, 1--9.

\bibitem[{Tan, Tombari, and Navab(2018)}]{tan2018real}
Tan, D.~J.; Tombari, F.; and Navab, N. 2018.
\newblock Real-time accurate 3d head tracking and pose estimation with consumer
  rgb-d cameras.
\newblock \emph{International Journal of Computer Vision}, 126: 158--183.

\bibitem[{Tian, Ang, and Lee(2020)}]{tian2020shape}
Tian, M.; Ang, M.~H.; and Lee, G.~H. 2020.
\newblock Shape prior deformation for categorical 6d object pose and size
  estimation.
\newblock In \emph{Computer Vision--ECCV 2020: 16th European Conference,
  Glasgow, UK, August 23--28, 2020, Proceedings, Part XXI 16}, 530--546.
  Springer.

\bibitem[{Ultralytics(2023)}]{Ultralyticsyolov5}
Ultralytics. 2023.
\newblock Yolov5: Real-time object detection.
\newblock \url{https://github.com/shanice-l/gdrnpp_bop2022}.

\bibitem[{Vince(2008)}]{vince2008geometric}
Vince, J. 2008.
\newblock \emph{Geometric algebra for computer graphics}.
\newblock Springer Science \& Business Media.

\bibitem[{Wang et~al.(2021)Wang, Manhardt, Tombari, and Ji}]{wang2021gdr}
Wang, G.; Manhardt, F.; Tombari, F.; and Ji, X. 2021.
\newblock Gdr-net: Geometry-guided direct regression network for monocular 6d
  object pose estimation.
\newblock In \emph{Proceedings of the IEEE/CVF Conference on Computer Vision
  and Pattern Recognition}, 16611--16621.

\bibitem[{Wang et~al.(2019)Wang, Sridhar, Huang, Valentin, Song, and
  Guibas}]{wang2019normalized}
Wang, H.; Sridhar, S.; Huang, J.; Valentin, J.; Song, S.; and Guibas, L.~J.
  2019.
\newblock Normalized object coordinate space for category-level 6d object pose
  and size estimation.
\newblock In \emph{Proceedings of the IEEE/CVF Conference on Computer Vision
  and Pattern Recognition}, 2642--2651.

\bibitem[{Wang, Chen, and Dou(2021)}]{wang2021category}
Wang, J.; Chen, K.; and Dou, Q. 2021.
\newblock Category-level 6d object pose estimation via cascaded relation and
  recurrent reconstruction networks.
\newblock In \emph{2021 IEEE/RSJ International Conference on Intelligent Robots
  and Systems (IROS)}, 4807--4814. IEEE.

\bibitem[{Weyrich et~al.(2007)Weyrich, Heinzle, Aila, Fasnacht, Oetiker,
  Botsch, Flaig, Mall, Rohrer, Felber et~al.}]{weyrich2007hardware}
Weyrich, T.; Heinzle, S.; Aila, T.; Fasnacht, D.~B.; Oetiker, S.; Botsch, M.;
  Flaig, C.; Mall, S.; Rohrer, K.; Felber, N.; et~al. 2007.
\newblock A hardware architecture for surface splatting.
\newblock \emph{ACM Transactions on Graphics (TOG)}, 26(3): 90--es.

\bibitem[{Xiang et~al.(2017)Xiang, Schmidt, Narayanan, and
  Fox}]{xiang2017posecnn}
Xiang, Y.; Schmidt, T.; Narayanan, V.; and Fox, D. 2017.
\newblock Posecnn: A convolutional neural network for 6d object pose estimation
  in cluttered scenes.
\newblock \emph{arXiv preprint arXiv:1711.00199}.

\bibitem[{Xiong et~al.(2023)Xiong, Muttukuru, Upadhyay, Chari, and
  Kadambi}]{xiong2023sparsegs}
Xiong, H.; Muttukuru, S.; Upadhyay, R.; Chari, P.; and Kadambi, A. 2023.
\newblock Sparsegs: Real-time 360 $\{$$\backslash$deg$\}$ sparse view synthesis
  using gaussian splatting.
\newblock \emph{arXiv preprint arXiv:2312.00206}.

\bibitem[{Zakharov, Shugurov, and Ilic(2019)}]{zakharov2019dpod}
Zakharov, S.; Shugurov, I.; and Ilic, S. 2019.
\newblock Dpod: 6d pose object detector and refiner.
\newblock In \emph{Proceedings of the IEEE/CVF international conference on
  computer vision}, 1941--1950.

\bibitem[{Zwicker et~al.(2001)Zwicker, Pfister, Van~Baar, and
  Gross}]{zwicker2001ewa}
Zwicker, M.; Pfister, H.; Van~Baar, J.; and Gross, M. 2001.
\newblock EWA volume splatting.
\newblock In \emph{Proceedings Visualization, 2001. VIS'01.}, 29--538. IEEE.

\bibitem[{Zwicker et~al.(2004)Zwicker, Rasanen, Botsch, Dachsbacher, and
  Pauly}]{zwicker2004perspective}
Zwicker, M.; Rasanen, J.; Botsch, M.; Dachsbacher, C.; and Pauly, M. 2004.
\newblock Perspective accurate splatting.
\newblock In \emph{Proceedings-Graphics Interface}, 247--254.

\end{thebibliography}

\end{document}